\documentclass{article}
\usepackage[utf8]{inputenc}

\usepackage{amsfonts}
\usepackage{amsmath}

\usepackage{graphicx}
\graphicspath{ {./images/} }

\usepackage{booktabs}

\usepackage{biblatex}
\usepackage[affil-it]{authblk}

\title{Machine learning reveals how personalized climate communication can both succeed and backfire}

\date{\today}

\author[1,*]{Harinen, Totte}
\author[1]{Filipowicz, Alexandre}
\author[1]{Hakimi, Shabnam}
\author[1]{Iliev, Rumen}
\author[1]{Klenk, Matthew}
\author[1]{Sumner, Emily Sarah}
\affil[1]{Toyota Research Institute}
\affil[*]{Corresponding author: totte.harinen@tri.global}

\begin{document}

\maketitle

\begin{abstract}
Different advertising messages work for different people. Machine learning can be an effective way to personalise climate communications. In this paper we use machine learning to reanalyse findings from a recent study, showing that online advertisements increased some people's belief in climate change while resulting in decreased belief in others. In particular, we show that the effect of the advertisements could change depending on people's age and ethnicity.
\end{abstract}

\section{Introduction}
Online advertising can play a major role in influencing beliefs around climate change. However, people are different and advertisements do not have the same effect on everyone. Sometimes advertisements can have the reverse effect of what they were initially intended for \cite{Swire-Thompson2020-kt}. For example, campaigns designed to decrease vaccine hesitancy could have the unintended consequence of making some segments of the population \textit{less} likely to get vaccinated \cite{Nyhan2014-no}. These are called \emph{backfire effects}. Consequently, an increasing body of research in machine learning focuses on the question of how to distinguish those for whom an advertising campaign works from those for whom it might backfire \cite{rzepakowski2012decision, Athey2016-lu, Wager2018-op, Li2019-yi, Gutierrez2017-kl, ascarza2018retention, Zhao2017-xd, Zhao2019-xr, Hansotia2002-ul, radcliffe2011real}. If we could predict how people would respond to certain messages, advertisements and communications about climate change could be displayed strategically to the right populations.

In this paper, we reanalyse a field experiment that uses online advertising to influence people's views about climate change \cite{Goldberg2021-lc}. This United States advertising campaign was specifically designed to influence Republicans' beliefs, since this group has historically expressed skepticism that climate change is indeed occurring. The messages in the campaign were framed around the influence of climate change on topics generally deemed important by Republicans (e.g. faith, security and immigration). The randomized experiment found an overall positive treatment effect in the group who were targeted with the advertisements. Given that the campaign was specifically designed for Republicans, the authors analyzed the effect by party affiliation. In this analysis, they found that, as expected, the effect was particularly strong for Republicans. For other political affiliations, the authors found either a smaller effect or no effect at all.

The results from this study provide compelling evidence that well-targeted ad campaigns can have a strong impact on target populations, including populations for which beliefs are traditionally hard to change, while having little to no negative consequences for other populations. However, by focusing their analysis exclusively on party affiliation, the authors miss the opportunity to explore the influence of their ad campaign on other population characteristics, and their potential interactions. Indeed, recent advances in machine learning-based targeted advertisement provide methods to assess the impact of advertisement on many, sometimes complex population segments. These advancements raise an important question for personalized climate-based communication strategies: instead of focusing on one population characteristic, can state of the art machine learning methods provide a deeper understanding of who to target--and who \textit{not} to target--with personalized climate based communication? To answer this question, we reanalyse the data reported by \cite{Goldberg2021-lc} by using machine learning-based targeting methods. Instead of focusing exclusively on party affiliation, we use all of the individual characteristics measured in the experiment to predict how treatment effects vary between groups. Although analyses of previous experiments have demonstrated how the effect of climate change messaging varies along single dimensions such as geography \cite{Zhang2018-dw}, our analysis is the first one to investigate such effect heterogeneity using multiple respondent characteristics simultaneously.

Our analysis of more complex population segments shows that, similar to the original paper, there are segments of the targeted populations for which targeted climate communication strategies worked exceptionally well--in some cases with effect sizes that far exceeded the average treatment effects originally reported by \cite{Goldberg2021-lc}. However, our results also show that these same ad campaigns reliably backfire for other segments of the same population, leading some people to believe less in climate change. Despite these observed backfire effects, we share the optimism that targeted ad campaigns can help build more effective communication strategies. We build on this optimism by proposing more modern advertisement analysis techniques that create a more holistic understanding of when, and to what extent targeted advertising campaigns lead to positive results, and identify when and for whom these campaigns could backfire.

\section{Findings}
\begin{figure}[t]
    \centering
    \includegraphics[width=0.8\textwidth]{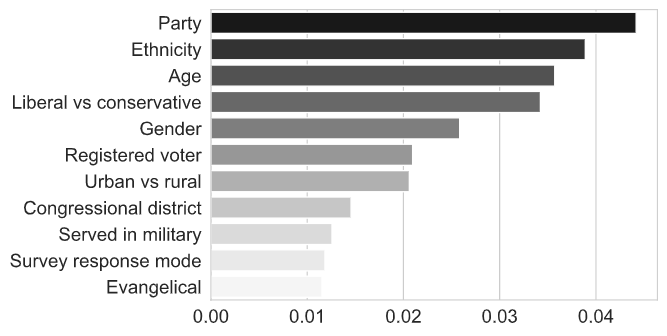}
    \caption{Feature importance scores for each of the 11 variables that we use to predict conditional average treatment effects. We use all of the covariates that were collected in the original study. The results show that party affiliation, ethnicity and age are the three variables most predictive of treatment effect heterogeneity.}
    \label{fig:feature_importances}
\end{figure}

We first reproduced the findings from \cite{Goldberg2021-lc}. As reported in their article, on average, Republicans assigned to the targeted ads increased their belief that climate change was happening by 7 percentage points, whereas beliefs for people with other political affiliations changed less. Having reproduced this overall main effect of targeted ads on Republicans, we next applied a machine learning approach to analyze the effect of the ads on larger demographic groups. The treatment effect of interest is the difference in the belief that climate change is happening between individuals in the treatment group and individuals in the control group. By a ``positive'' treatment effect we refer to an increase in the belief and by a ``negative'' treatment effect we refer to a decrease in the belief. Our goal is to predict how this treatment effect changes for different groups of individuals. As predictors, we use 11 variables that are related to the study participants' demographic characteristics (the full list of variables is shown in the Y-axis of Figure \ref{fig:feature_importances}).

In machine learning, the standard practice for evaluating predictive performance is to split the dataset randomly into training and testing samples. The model learns its parameters on the training set and makes predictions on the testing set. Here, we follow the same practice except for one difference: whereas standard machine learning models usually try to predict outcomes such as whether or not a subject belongs to a class, our model instead predicts the effect of the treatment for different groups of individuals. To illustrate, the machine learning algorithm might consider the difference in the effect of the advertisement campaign for those who are under 45 years old and those who are over 45 years old. The model considers multiple complex combinations of demographic characteristics such as age in order to identify the groups of individuals with the most negative and most positive treatment effects. More details on the model can be found in Section \ref{sec:methods}.

\begin{figure}[t]
    \centering
    \includegraphics[width=0.75\textwidth]{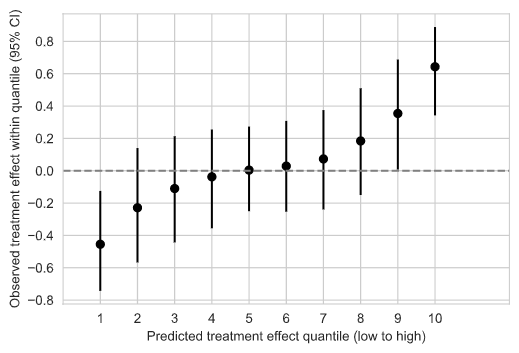}
    \caption{Observed treatment effects in testing samples sorted according to predicted treatment effects. For each quantile, the distribution of observed treatment effects is obtained by making predictions using 1000 bootstrapped datasets. The figure shows a clear correlation between predicted and observed treatment effects, indicating that our machine learning model is able to distinguish those individuals who are likely to respond to the advertisement campaign negatively from those who are likely to respond positively.}
    \label{fig:subgroup_ates}
\end{figure}

The predicted treatment effects and their 95\% empirical confidence intervals are shown in Figure \ref{fig:subgroup_ates}. The results show that our model can reliably predict the individuals who have the most negative and most positive treatment effects in the experiment. Moreover, we observe very large average treatment effects within the groups with the most negative and the most positive predicted changes in their beliefs. For the 10\% of the sample who we predict to have the largest negative change, the belief that climate change is happening is on average 45 percentage points \textit{less} in the treatment group that it is in the control group (who received none of the online advertisements). For the 10\% of sample with the most positive predicted reaction, the average increase in the belief that climate change is happening is 64 percentage points for those in the treatment group. This shows that the exact same campaign can have either a highly positive or a highly negative effect depending on who you target.

Having established that we can predict the individuals with the most negative and most positive treatment effects, we next examine which study participant characteristics predict these differences. To do so, we use another machine learning model to predict the \textit{output} of our treatment effect heterogeneity model. As predictors in this meta-level model, we use the same 11 study participant characteristics as we did in the original treatment effect heterogeneity analysis. After fitting the model, we examine its feature importance scores, which are a common way to evaluate the contribution of predictors in machine learning. The results of this analysis are shown in Figure \ref{fig:feature_importances}. Here, we can see that the features that are most predictive of treatment effects include party affiliation, ethnicity, age, political ideology and sex. Notably, only party affiliation was considered as an explanatory variable in the original analysis \cite{Goldberg2021-lc}.

\begin{figure}
    \centering
    \includegraphics[width=.75\textwidth]{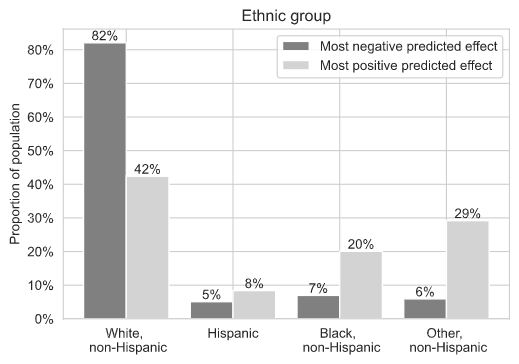}
    \includegraphics[width=.75\textwidth]{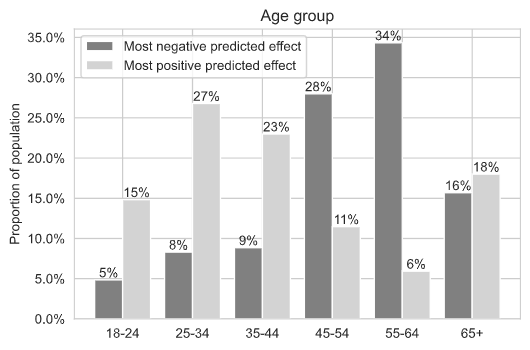}
    \caption{Ethnicity and age characteristics by predicted response to the advertising campaign. The comparison is between the 10\% of the sample with the most negative predicted response and the 10\% of the sample with the most positive predicted response. The proportions represent the averages in 1000 bootstrapped samples. The figure shows that those with the most negative predicted response tend to be white and middle-aged whereas those with the most positive predicted response tend to be non-white and younger.}
    \label{fig:demog_x_bucket}
\end{figure}

Importance scores do not in and of themselves tell us how features are related with treatment effects. However, we can examine how the characteristics of those with the highest negative effect differ from those with the highest positive effect. By doing this, we can understand the directionality of the relationship better. We conduct this type of analysis for the two most important features besides party affiliation, which was already reported in the original study. The result of our analysis are show in Figure \ref{fig:demog_x_bucket}. These results indicate that the advertising campaign tends to backfire in populations who are white and middle-aged. The most positive treatment effects, by contrast, are predicted to happen in younger, non-white populations.

\begin{figure}
    \centering
    \includegraphics[width=.75\textwidth]{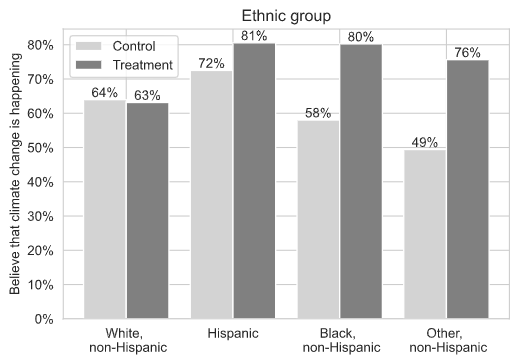}
    \includegraphics[width=.75\textwidth]{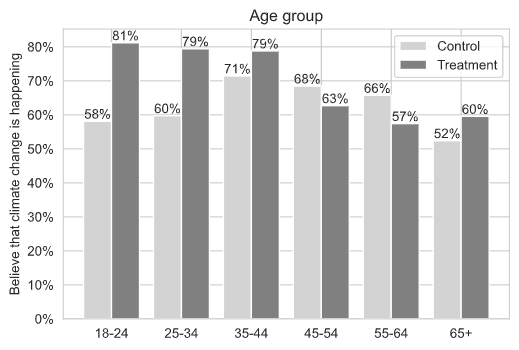}
    \caption{The relationship between the treatment effect and two demographic characteristics, ethnicity and age. The figures show that the relationships suggested by our machine learning model also hold if we examine the experimental population as a whole. In particular, the positive difference between treatment and control tends to occur in younger non-white populations.}
    \label{fig:demog_x_ate}
\end{figure}

The demographic differences that we observe by comparing the groups with the most negative and most positive predicted treatment effects are only in rare occasions visible when we observe the entire population. This is because the differences tend to get diluted when we include the rest of the sample in our comparison. However, given the very pronounced differences in demographic characteristics that we find in our machine learning based analysis, we conduct a follow up investigation to examine the relationship between the treatment condition, ethnicity and age in the experimental population as a whole. For this analysis, we use a standard multiple regression approach. Interestingly, the population-level multiple regression model also indicates that age and ethnicity interact with the treatment effect (see Section \ref{sec:methods} for details). Figure \ref{fig:demog_x_ate} show the difference between the treatment and control condition for different age groups and ethnicities. These results show that the differences observed by examining those with the most positive and most negative predicted responses are so strong that they are also visible at the population level.

\section{Discussion}
Our results show that communications about climate change can influence people in different ways, even backfiring in certain populations. In the case of the advertising campaign examined above, we found that the communications worked well for non-white younger populations and  backfired for white, middle-aged populations. These findings have important practical implications. First, if the advertising campaign analysed here were to be introduced to a larger population, it would make sense to target younger non-white audiences to maximize return of investment. Second, to maximize the effect of climate change information campaigns in general, we recommend adopting a strategy where ads are first piloted in randomized experiments at a smaller scale. A targeting model can then be trained on the treatment effects observed in the pilot experiments, so that in the full campaign the advertiser can selectively target those populations that are predicted to have the desired change in belief, and avoid targeting those populations for whom the campaign is predicted to backfire \cite{ascarza2018retention}. We agree with \cite{Goldberg2021-lc} that online advertising can play an important role in shifting views about climate change, and we believe this is particularly true if such advertising is combined with modern causal machine learning tools that help us to ensure that communications will not backfire.

\section{Methods}\label{sec:methods}
The outcome variable in our analysis is people's answer to the question ``Do you think that global warming is happening?'' The possible answers to this question were ``Yes'', ``Don't know" and ``No''. Following the original analysis in \cite{Goldberg2021-lc}, we binarize the outcome such that ``Yes'' is recoded into ``1'' and the other two possible answers are recoded into ``0''. The demographic characteristics that we use to predict treatment effects are shown in Figure \ref{fig:feature_importances}. Because the variables we use for prediction can be considered as categorical, we use the one-hot encoding approach in which each value of a variable (except for one) are represented as separate binary variables. We use these binary variables rather than the original features as our predictors.

We use machine learning to estimate heterogeneity in treatment effects. If $Y$ is the outcome of interest and $W$ is the treatment group assignment, and both $Y$ and $W$ have possible values in ${0, 1}$, we can define the average treatment effect (ATE) as
\begin{equation}
    \mathbb{E}[Y = 1 \mid do(W = 1)] - \mathbb{E}[Y = 1 \mid do(W = 0)]
\end{equation}
where $do(W = w)$ denotes \textit{setting} as opposed to \textit{observing} the value of the treatment condition \cite{Pearl2009-gh, pearl2016causal}. Randomly assigning a treatment group is one example of an event denoted by the $do$-operator. Suppose we conduct an experiment where participants can freely choose whether they are in the treatment or control group. In this case, we merely observe the treatment assignment and write $W = w$. Now suppose we conduct an experiment where we randomly allocate participants into treatment and control groups. In this case, we actively intervene in the process that generates the data and write $do(W = w)$. One way to appreciate the usefulness of the $do$-operator is to note that the treatment effects observed in these two hypothetical experiments would likely be quite different.

As the results of this paper show, a positive ATE does not guarantee that the treatment did not backfire for some populations. To find populations with different treatment effects, we consider the conditional average treatment effect (CATE), which is defined as
\begin{equation}
\begin{aligned}
    \tau = & \mathbb{E}[Y = 1 \mid do(W = 1), X = x] - \\
    & \mathbb{E}[Y = 1 \mid do(W = 0), X = x]
\end{aligned}
\end{equation}
where $X$ represents some vector of covariates, such as the demographic characteristics of a participant in a study. Figure \ref{fig:demog_x_ate} shows CATEs for different ethnic and age groups.

There is an increasing literature on how to estimate CATEs effectively using machine learning methods \cite{Kunzel2019-ma}. The method used here, known as T-learner, is one of the simplest and most studied approach for CATE learning. T-learner is based on estimating two separate response functions: one for the outcome under the control condition and one for the outcome under the treatment condition. Hence, the ``T'' in T-Learner stands for ``two''. Formally, the response under control is defined as
\begin{equation}
    \mu_0 = \mathbb{E}[Y = 1 \mid do(W = 0), X = x]
\end{equation}
and the response under treatment is defined as
\begin{equation}
    \mu_1 = \mathbb{E}[Y = 1 \mid do(W = 1), X = x]
\end{equation}
The T-learner estimates $\mu_0$ by predicting $Y$ as a function of $X$ using the control observations only; and it estimates $\mu_1$ by predicting $Y$ as a function of $X$ using the treatment observations only. The difference between these estimates is then taken as the estimated CATE:
\begin{equation}
    \hat{\tau} = \hat{\mu_1} - \hat{\mu_0}
\end{equation}
The model we use to estimate the response functions is the gradient boosting algorithm in the scikit-learn machine learning library. \cite{Friedman2001-cb, pedregosa2011scikit} We use default hyperparameters in all of our models.

To evaluate the performance of the T-learner, we use a bootstrapping approach \cite{Efron1979-kk}. We form 1000 new populations of size $N = 1600$ sampled (with replacement) from the original population. We divide each population into training and testing sets using a 80\%/20\% split. For each population, we train the T-learner into the training set and make predictions of $\hat{\tau}$ on the testing set. The testing set is then divided into 10 quantiles according to the predicted $\hat{\tau}$. We use 10 quantiles so as to ensure a sufficient number of treatment and control observations within each group. We calculate the \textit{observed} difference between those in the treatment group and those in the control group within each quantile. We collect these observed differences across all 1000 bootstrap samples and calculate the treatment effects as well as their empirical 95\% confidence intervals for each 10 quantiles. These metrics are shown in Figure \ref{fig:subgroup_ates}. To calculate the feature importances, we model $\hat{\tau}$ as a function of $X$ in each of the 1000 bootstrap samples using a separate gradient boosting regressor. We then calculate the Gini feature importances \cite{Menze2009-xt} for each of these 1000 gradient boosting regressors and average the importance scores. Further, because the features are one-hot encoded, we average the importance scores across the different categories of each feature.

Finally, we use ordinary least squares regression \cite{Gomila2021-fe} to examine the relationships between the treatment condition, age and ethnicity. The details of these analyses are shown in Tables \ref{table:eth_ols} and \ref{table:age_ols}.

\begin{table}
\begin{center}
\begin{tabular}{lcccccc}
\toprule
                                  & \textbf{coef} & \textbf{std err} & \textbf{t} & \textbf{P$> |$t$|$} \\
\midrule
\textbf{Intercept}              &       0.6397  &        0.019     &    32.810  &         0.000   \\
\textbf{Ethnicity[Hispanic]}    &       0.0853  &        0.078     &     1.098  &         0.272   \\
\textbf{Ethnicity[Black]}       &      -0.0595  &        0.056     &    -1.057  &         0.291   \\
\textbf{Ethnicity[Other]}       &      -0.1456  &        0.055     &    -2.642  &         0.008   \\
\textbf{Condition}              &      -0.0083  &        0.027     &    -0.303  &         0.762   \\
\textbf{Condition:Ethnicity[Hispanic]} &    0.0889  &        0.113     &     0.790  &         0.430   \\
\textbf{Condition:Ethnicity[Black]} &       0.2305  &        0.080     &     2.898  &         0.004   \\
\textbf{Condition:Ethnicity[Other]} &       0.2706  &        0.079     &     3.408  &         0.001   \\
\bottomrule
\end{tabular}

\begin{tabular}{lclc}
\textbf{R-squared:}     &   0.018 & \textbf{F-statistic:}   &   4.152 \\
\textbf{Df Residuals:}  &   1592 \\
\bottomrule
\end{tabular}
\end{center}
\caption{OLS regression results for the relationship between ethnicity and the treatment condition. The dependent variable is belief in climate change. The contrast category for Ethnicity is is ``White''.}
\label{table:eth_ols}
\end{table}

\begin{table}
\begin{center}
\begin{tabular}{lcccccc}
\toprule
                            & \textbf{coef} & \textbf{std err} & \textbf{t} & \textbf{P$> |$t$|$} \\
\midrule
\textbf{Intercept}            &       0.5811  &        0.055     &    10.592  &         0.000       \\ 
\textbf{Age[25-34]}           &       0.0163  &        0.077     &     0.212  &         0.832       \\ 
\textbf{Age[35-44]}           &       0.1332  &        0.071     &     1.884  &         0.060       \\ 
\textbf{Age[45-54]}           &       0.1035  &        0.067     &     1.542  &         0.123       \\ 
\textbf{Age[55-64]}           &       0.0762  &        0.065     &     1.168  &         0.243       \\
\textbf{Age[65+]}             &      -0.0573  &        0.064     &    -0.898  &         0.369       \\
\textbf{Condition}            &       0.2307  &        0.075     &     3.074  &         0.002       \\ 
\textbf{Condition:Age[25-34]} &      -0.0340  &        0.103     &    -0.328  &         0.743       \\ 
\textbf{Condition:Age[35-44]} &      -0.1571  &        0.099     &    -1.581  &         0.114       \\ 
\textbf{Condition:Age[45-54]} &      -0.2884  &        0.094     &    -3.076  &         0.002       \\
\textbf{Condition:Age[55-64]} &      -0.3138  &        0.091     &    -3.441  &         0.001       \\
\textbf{Condition:Age[65+]}   &      -0.1589  &        0.088     &    -1.814  &         0.070       \\
\bottomrule
\end{tabular}
\begin{tabular}{lclc}
\textbf{R-squared:}     &   0.034  &    \textbf{F-statistic:}   &   5.054 \\
\textbf{Df Residuals:}  &   1588 \\
\bottomrule
\end{tabular}
\end{center}
\caption{OLS regression results for the interaction between age and the treatment condition. The dependent variable is belief in climate change. The contrast category for Age is ``18-24''.}
\label{table:age_ols}
\end{table}

\newpage
\printbibliography

\end{document}